\documentclass{article}
\PassOptionsToPackage{numbers, compress}{natbib}

\usepackage[final]{neurips_2021}
\usepackage{xr-hyper}

\usepackage[utf8]{inputenc} 
\usepackage[T1]{fontenc}    
\usepackage{hyperref}       
\usepackage{url}            
\usepackage{booktabs}       
\usepackage{amsfonts}       
\usepackage{nicefrac}       
\usepackage{microtype}      
\usepackage[dvipsnames]{xcolor}         
\usepackage{siunitx}        

\usepackage{todonotes}
\usepackage{subcaption}

\usepackage{amsmath}
\usepackage{mathtools}  
\usepackage{amsfonts}

\newcommand{\MDP}[0]{\mathcal{M}}                       
\newcommand{\statespace}[0]{\mathcal{S}}                
\newcommand{\actionspace}[0]{\mathcal{A}}               
\newcommand{\transdomain}[0]{\mathcal{T}}               
\newcommand{\rewards}[0]{\mathcal{R}}                   

\newcommand{\inst}[0]{i}                    
\newcommand{\insts}[0]{\mathcal{I}}         

\newcommand{\cMDP}[0]{\MDP_\insts}                      
\newcommand{\cMDPdef}[0]{\cMDP \coloneqq \{\MDPi\}_{\inst \sim \insts}} 
\newcommand{\MDPi}[0]{\MDP_\inst}                       
\newcommand{\MDPidef}[0]{\MDPi \coloneqq (\statespace, \actionspace, \transdomaini, \rewardsi)}  

\newcommand{\transdomaini}[0]{\transdomain_\inst}       
\newcommand{\rewardsi}[0]{\rewards_\inst}               

\usepackage{wrapfig}
\usepackage{xspace}
\usepackage{pifont}
\usepackage{xcolor}
\usepackage[inline,shortlabels]{enumitem}

\makeatletter
\newcommand*{\addFileDependency}[1]{
  \typeout{(#1)}
  \@addtofilelist{#1}
  \IfFileExists{#1}{}{\typeout{No file #1.}}
}
\makeatother

\title{\centering
Hyperparameters in Contextual RL \\ are Highly Situational}
\author{%
Theresa Eimer\thanks{Equal Contribution, Contact Author}$^{\;\:}$, Carolin Benjamins$^{\tiny{*}}$ and \textbf{Marius Lindauer}\\
Leibniz University Hanover\\
\texttt{$\left\{\right.$t.eimer,c.benjamins$\left.\right\}$@ai.uni-hannover.de} \
}

\begin{document}

\maketitle

\begin{abstract}
Although Reinforcement Learning (RL) has shown impressive results in games and simulation, real-world application of RL suffers from its instability under changing environment conditions and hyperparameters. We give a first impression of the extent of this instability by showing that the hyperparameters found by automatic hyperparameter optimization (HPO) methods are not only dependent on the problem at hand, but even on how well the state describes the environment dynamics. Specifically, we show that agents in contextual RL require different hyperparameters if they are shown how environmental factors change. In addition, finding adequate hyperparameter configurations is not equally easy for both settings, further highlighting the need for research into how hyperparameters influence learning and generalization in RL.
\end{abstract}
\section{Introduction}
Even though reinforcement learning (RL) has shown considerable progress in many areas like game playing \citep{silver-nature16a,badia-icml20}, robot manipulation \citep{lee-sciro20}, traffic control~\citep{arel-its10a}, chemistry~\citep{zhou-acs17a} and logistics~\cite{li-aamas19a}, deploying RL in application remains challenging.
This is especially in high-stakes domains like autonomous driving and healthcare where failures can be fatal.
One explanation is that the design of modern RL agents does not prioritize generalization, making them susceptible to even small variations in their environment or hyperparameters \citep{henderson-aaai18a,islam-corr17,lu-amia20,meng-data19}. 

Different alterations of an environment and the impact on the agents' performance can be modelled and studied e.g. via contextual RL.
For instance, in OpenAI's pendulum environment \cite{gym} the task is to exert force upon a pendulum such that it balances upright.
The other factors defining this process, often physical attributes like the length of the balancing pole, its mass or even the magnitude of gravity, can be chosen as context features. By varying context features during training and testing we are able to represent different instances of the same environment, see Figure~\ref{fig:concept_pendulum}. 
This will challenge the agent to perform well across environmental changes and ultimately gives a better estimate of its robustness and generalization capabilities, a step closer to reliable real-world application of RL~\cite{benjamins-arxiv21}.

While cRL opens many new research directions, it also introduces more variables into an already brittle RL pipeline. 
From the perspective of a RL practitioner, this can further complicate the process of finding an agent that solves a given task - and indeed we easily show it does.
Our research hypotheses are:
\begin{enumerate}
    \item Hyperparameter configurations for RL agents in the contextual setting are very sensitive to even small changes in the task.
    \item Well-performing hyperparameter configurations for RL agents depend on whether context features are explicitly provided or not.
\end{enumerate}

We study validity of these research hypotheses in our experiments on several contextually extended environments from the benchmark library CARL~\citep{benjamins-arxiv21}. 
Our results demonstrate the importance of developing and applying AutoRL methods to tune existing RL methods in accordance with their task as well as generating better insights into where and why RL agents fail.



\section{Related Work}
The sensitivity of RL algorithms to changes in their hyperparameters~\cite{henderson-aaai18a, islam-corr17} has been a first indicator for the need of hyperparameter optimization.
Research into which hyperparameters and algorithm components are deciding factors for different classes of algorithms like policy gradient \cite{engstrom-iclr20, andrychowicz-iclr21} or off-policy algorithms \cite{obando-icml21} has contributed to our fundamental understanding of RL algorithms, even though it has not solved their instability issues.

Automatically learning or tuning RL pipelines for a given problem can instead significantly boost performance. There is a broad span of methods, from learning RL algorithms from scratch \cite{wang-corr16,coreyes-iclr21}, to learning algorithm components \cite{finn-icml17a,duan-corr16} or tuning the agent's hyperparameters \cite{jaderberg-arxiv17a, hertel-corr20,parkerholder-neurips20, franke-iclr21}.

While prior work on hyperparameters in RL shows that RL algorithms are sensitive to hyperparameters and greatly benefit from optimized hyperparameters settings, aptitudes of hyperparameter configurations on changing or varying training environments are underexplored.
When the goal is robust RL, however, it is just as important to determine how sensitive the hyperparameter configuration is to environment perturbations as to find a well-performing configuration.

\section{Contextual Reinforcement Learning}
\label{sec:cRL_def}
A contextual Markov Decision Process (cMDP)~\citep{hallak-corr15,modi-alt18} is defined as a set of multiple MDPs $\cMDPdef$ characterized by the instance $\inst$ sampled from the instance set or distribution $\insts$.
The MDP is a 4-tuple $\MDPidef$ consisting of a state space $\statespace$, action space $\actionspace$, transition function $\transdomaini$ and reward function $\rewardsi$. 
The transition function $\transdomaini$ and reward function $\rewardsi$ are subject to change across instances.
With this formulation we can express slight variations in the environment, e.g., varying lengths and masses of a pendulum, and therefore train for generalization.
The objective can vary according to the current application, e.g. maximizing the expected return across a test instance set or for 
a single, hard instance.

\begin{wrapfigure}{r}{0.38\linewidth}
    \centering
    \vspace{-1em}
    \includegraphics[width=0.3\textwidth]{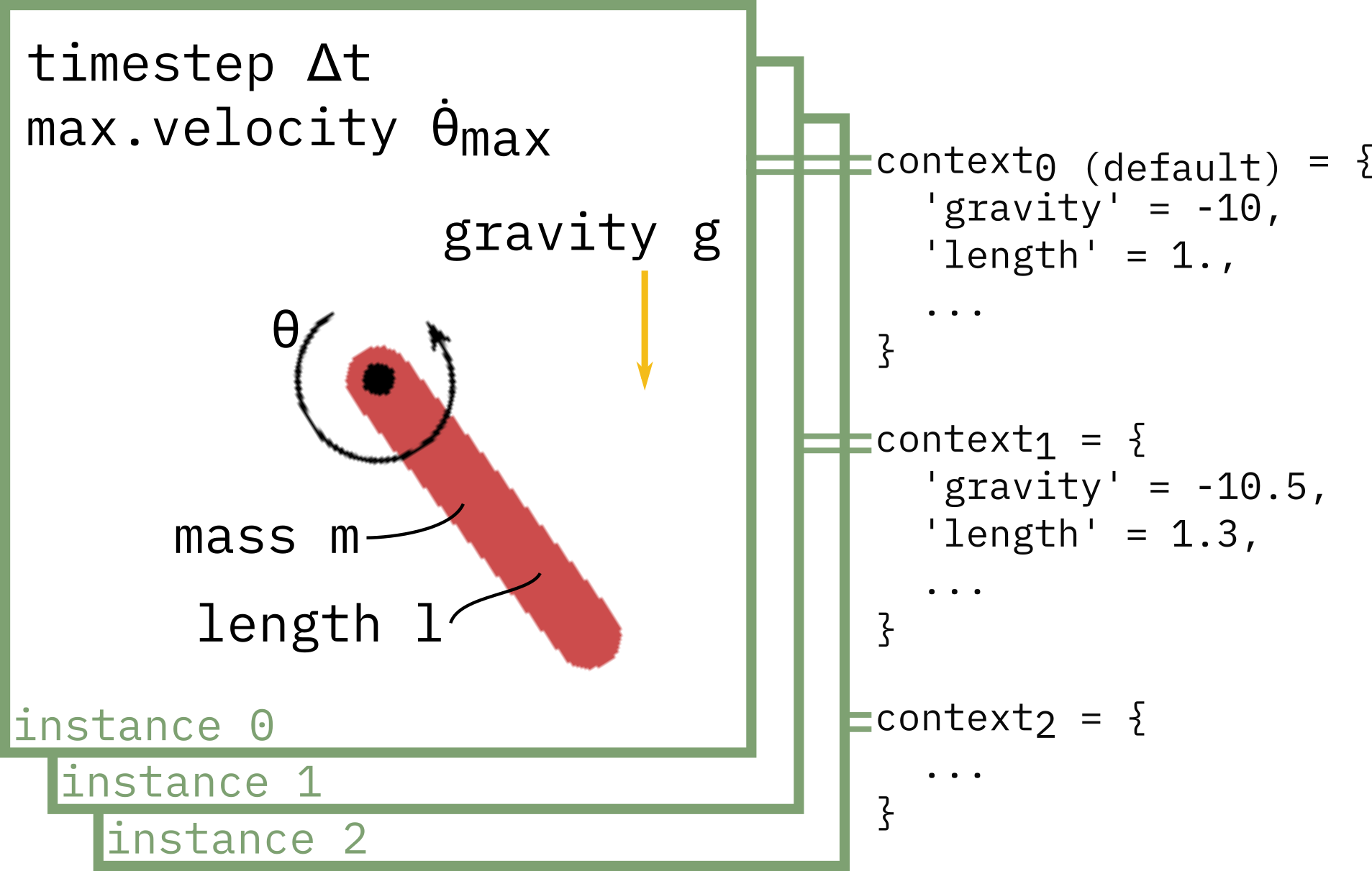}
    \caption{In the contextual RL setting Pendulum's~\cite{gym} physical parameters are varied across instances while the underlying dynamic equations stay the same.}
    \label{fig:concept_pendulum}
\end{wrapfigure}
In order to analyze the effect of HPs in this setting we use the CARL (Contextually Adaptive RL) benchmark library~\cite{benjamins-arxiv21}.
CARL extends well-known environments and makes the context defining the behavior of the environment configurable and optionally visible.
The context is often based on physical parameters like gravity or friction, see Figure~\ref{fig:concept_pendulum}.
In our experiments, we use these benchmarks to generalize over different instances (contexts) of the same environment, drawn from a common context distribution.
The benchmark includes environments from Open AI's gym~\cite{gym} (classic control and box2d), Google Brax'~\cite{brax2021github} locomotion environments as well as Super Mario (TOAD-GAN)~\cite{awiszus-aiide20,schubert-tg21} controlling level similarity and a RNA folding environment~\cite{runge-iclr19a}.


\section{Experiments}
We evaluate how context and changes in environments influence the meta-problem of setting the hyperparameters.
For this purpose, we use PB2~\cite{parkerholder-neurips20} to optimize the hyperparameters of a standard DDPG~\cite{lillicrap-iclr16} algorithm on CARLPendulum and PPO~\cite{schulman-arxiv17a} on CARLAcrobot and CARLLunarLander~\citep{benjamins-arxiv21}.
For all environments we first use 8 parallel PB2 workers on one seed\footnote{We note that one PB2 run with a single seed only allows us to draw preliminary conclusions from our results and further runs are needed to verify our findings. Nevertheless, we believe that our results provide an important first indication for potential challenges in AutoRL for cRL.} to optimize the hyperparameters during the training with a time budget of $\SI{24}{\hour}$ and $150\, \text{GB}$ of memory.
For DDPG, we optimize the learning rate, discount factor and soft update $\tau$; for PPO the learning rate, discount factor, entropy and value function coefficients, the maximum gradient normalization and GAE parameter. 
The optimization bounds of the hyperparameters can be found in Appendix~\ref{sec:hardware_software_HPs}.
For testing the hyperparameter policies we use 5 new seeds.
Please see the Appendix~\ref{sec:hardware_software_HPs} for further details on the hardware, software and hyperparameters.
The code for these experiments along with the found hyperparameter schedules is available at a anonymous repository during the review period and the actual one will be public upon acceptance: \mbox{\url{https://github.com/automl-private/cRL_HPO}} 

\subsection{Tunability of Visible and Hidden Context Features}
Using the setup from above we varied the gravity for CARLPendulum and the length of the first link for CARLAcrobot.
Similar trends can be observed for both environments.
First of all, visible context information in combination with hyperparameter optimization leads to better overall performance and even learning speed.
For CARLPendulum a final evaluation score of $-312$ was achieved for hidden context and $-262$ for visible context.
For CARLAcrobot the difference is more substantial: $-184.65$ for hidden, $-80.13$ for visible context, on the best seed, resp.

\begin{figure}
    \centering
    \begin{subfigure}[t]{0.49\textwidth}
        \centering
        \includegraphics[width=1\textwidth]{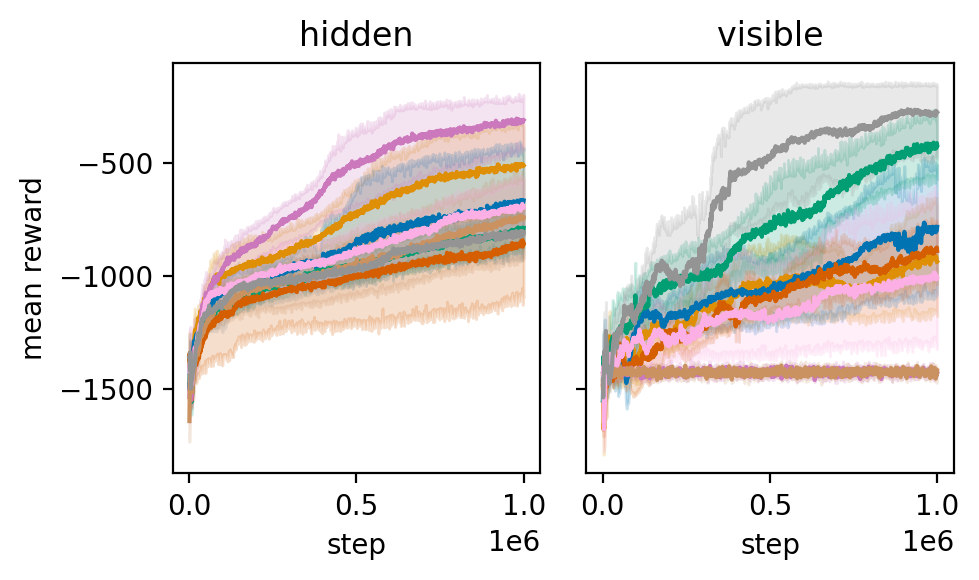}
        \caption{CARLPendulum}
        \label{fig:results_pb2_pendulum}
    \end{subfigure}
    \begin{subfigure}[t]{0.49\textwidth}
        \centering
        \includegraphics[width=1\textwidth]{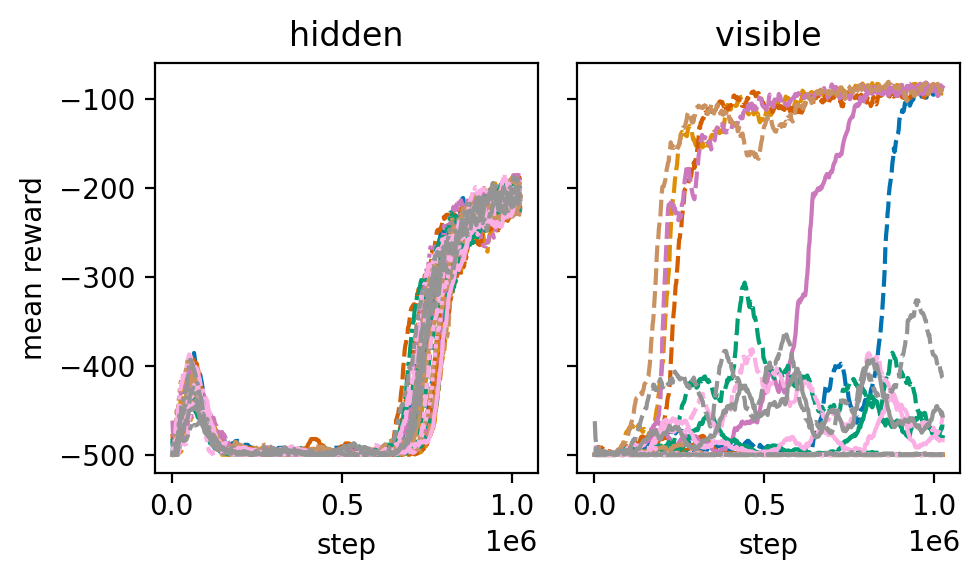}
        \caption{CARLAcrobot}
        \label{fig:results_pb2_acrobot}
    \end{subfigure}
    \caption{Training performance with hidden and visible context on each hyperparameter schedule found by PB2.The different line colors indicate schedules found by individual PB2 workers, left with confidence intervals, right with one line for each seed.}
\end{figure}

Our second observation is that hyperparameter optimization on visible context information seems to be much harder.
In particular on CARLAcrobot, only some seeds of the hyperparameter schedule evaluation are able to perform well and most of them fail.
There is not only a large performance gap between schedules, but also between different random seeds which we do not observe if we hide the context, see lines with same colors in Figure~\ref{fig:results_pb2_acrobot}.
On CARLPendulumn, there is also some performance spread for hidden context information, but the spread is wider for visible context information.

Last but not least, we also looked into some of the hyperparameter schedules found by PB2, see Figure~\ref{fig:pb2_hp_schedules} in the appendix.
These schedules change more for visible than for hidden context information. 
This is another indication that hyperparameter optimization for cRL with visible context information is harder, although more explicit information is provided.

\subsection{A Failure Case: LunarLander}
\begin{figure}[ht]
    \centering
    \includegraphics[width=0.6\textwidth]{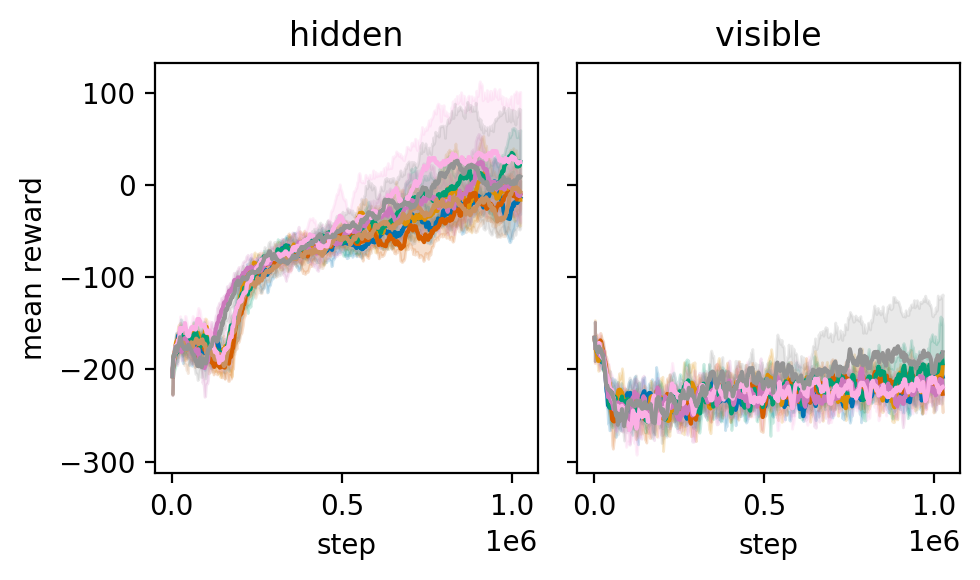}
    \caption{Training performance with hidden and visible context on each hyperparameter schedule found by PB2 on CARLLunarLander. The different line colors indicate schedules found by individual PB2 workers with confidence intervals.}
    \label{fig:results_LunarLander}
\end{figure}

With the same experimental setup as before, we optimzed the hyperparameters on the CARLLunarLander environment across varying gravity settings.
The results paint a different picture than above (see Figure~\ref{fig:results_LunarLander}): While PB2 was able to find hyperparameter schedules resulting in a positive performance trend for hidden contexts, it could not do so for visible contexts.
In addition, although the agent with the hidden contexts reaches higher rewards, it was still not able to solve the environment.

There are several possible reasons for these observations. We believe that the most likely ones include:
\begin{enumerate*}[(i)]
    \item The variation introduced into CARLLunarLander via the different contexts might be much larger or much harder to learn than in CARLPendulum and CARLAcrobot. In the case of hidden context information, the agent might not be able to sufficiently distinguish between different instances. The sub-optimal performance even in the case of hidden context information compared to the solved score of $200$ is an indication for that. In the case of visible context information, the agent is able to distinguish between the instances, but cannot properly learn to solve the environment potentially due to catastrophic forgetting, which cannot be compensated by HPO alone.
    \item The visible context information provides so much added information to process, that the capacity of the default policy network might be insufficient. Therefore the agent cannot learn when the context is visible, but improves at least to a degree when it is hidden. This would imply that neural architecture search~\cite{elsken-jmlr19a} would become much more important for RL. Some further preliminary results supporting this are shown in Appendix~\ref{app:sizes}.
    \item For each instance, a different optimal hyperparameter configuration is needed and there is no single hyperparameter configuration that performs well on all the tasks. In the related field of algorithm configuration this is a typical observation for some tasks~\cite{xu-aaai10a}.
    \item Because HPO is much harder in this case (for unknown reasons), the hyperparameter optimization could not move past a local optimum in the learning rate (see Appendix~\ref{sec:hardware_software_HPs}) for the visible case, resulting in a suboptimal hyperparameter configuration. This hypothesis would also be supported by the results on CARLPendulum and CARLAcrobot.
\end{enumerate*}

At the moment, we cannot say for certain which of these reasons caused our experiments to fail or if more factors are involved.
Our findings, however, are a further indication that the cRL setting, especially as environments grow more challenging, is harder to navigate for AutoRL methods compared to standard RL.

\section{Conclusion and Future Work}
For the contextual RL setting we show that tuning hyperparameters for the RL algorithm at hand plays a crucial role in solving the environment. 
Especially in the case where we provide the agent with the context, the hyperparameter configuration can decide between success and failure. 
Even so, using hyperparameter optimization in cRL is more complex than for standard RL tasks and as we have seen, finding well-performing configurations is not a guarantee.
The addition of context makes the learning process more difficult and harder to optimize for and finding the source of failure can be hindered by the fact that we have comparatively little knowledge about the cRL setting.
Thus there is a need for more insights into cRL in general and AutoRL for cRL in particular.

With these preliminary results we want to motivate further research both into how RL agents generalize in the first place and how we can more easily adapt their hyperparameters to support this goal. 
Our results also clearly show the importance of using existing AutoRL methods when applying RL to a task and the potential improvements gained.
Lastly, they emphasise that the methods we currently use for RL are constrained to their problem setting by factors including their hyperparameters.
If we want to move beyond single task settings and towards general RL, we need to refine our understanding of the interplay of RL algorithms and their hyperparameters and develop efficient AutoRL practices.
\section*{Acknowledgements}
The authors acknowledge funding by the German Research Foundation (DFG) under LI 2801/4-1.

\bibliographystyle{apalike}
\bibliography{bib/meta-gym,bib/strings,bib/lib,bib/proc,bib/tune_hps}

\appendix

\section{Hardware, Software and Hyperparameters}
\label{sec:hardware_software_HPs}
\paragraph{Hardware} All experiments are executed on a slurm GPU cluster consisting of six nodes with six to eight Nvidia RTX 2080 Ti GPUs each.

\paragraph{Software} Our implementation uses the agents from stablebaselines3~\cite{stable-baselines3}.

For CARLPendulum~\cite{benjamins-arxiv21} we tune the hyperparameters of the DDPG agent~\cite{lillicrap-iclr16}, for CARLAcrobot and CARLLunarLander we use the PPO agent~\cite{schulman-corr17}.
For tuning we use PB2~\cite{parkerholder-neurips20} as implemented in the ray tune package~\cite{ray-2019}.

Our experiments can be reproduced via the scripts we provide at \mbox{\url{https://github.com/automl-private/cRL_HPO}}.

\paragraph{PB2 Usage}
PB2~\cite{parkerholder-neurips20} runs with 8 workers with a total timelimit of $\SI{24}{\hour}$ and a memory limit of $150\,\text{GB}$.
After $4096$ environment steps the hyperparameter configurations are adjusted by PB2.
We start the optimization with a batch size of $128$, a learning rate of $0.00003$ and discount factor of $0.99$ for both algorithms. All other hyperparameters start at their default values.

In both cases, the learning rate is limited to be between $0.00001$ and $0.02$ and the discount factor between $0.8$ and $0.999$.
For DDPG~\cite{lillicrap-iclr16}, $\tau$ can lie between $0.0$ and $0.99$.
In PPO~\cite{schulman-corr17}, the maximum gradient normalization and value function coefficient are both limited to between $0.0$ and $1.0$, the entropy coefficient to between $0.0$ and $0.5$ and the GAE parameter is between $0.8$ and $0.999$.

\begin{figure}[ht]
    \centering
    \begin{subfigure}[t]{0.3\textwidth}
        \centering
        \includegraphics[width=1\textwidth]{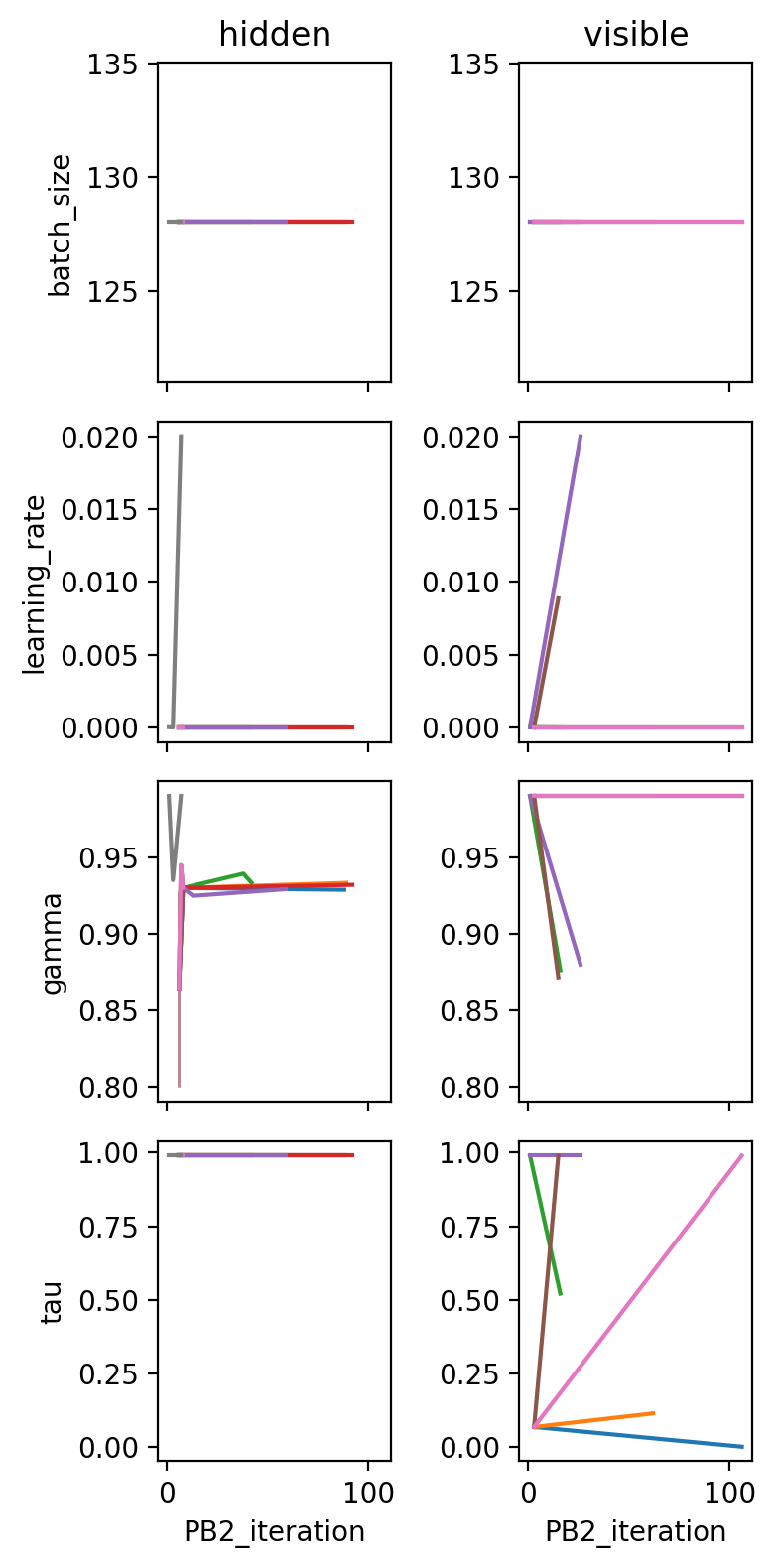}
        \label{fig:pb2_hp_schedule_CARLPendulum}
        \caption{CARLPendulum}
    \end{subfigure}
    \quad
    \begin{subfigure}[t]{0.3\textwidth}
        \centering
        \includegraphics[width=1\textwidth]{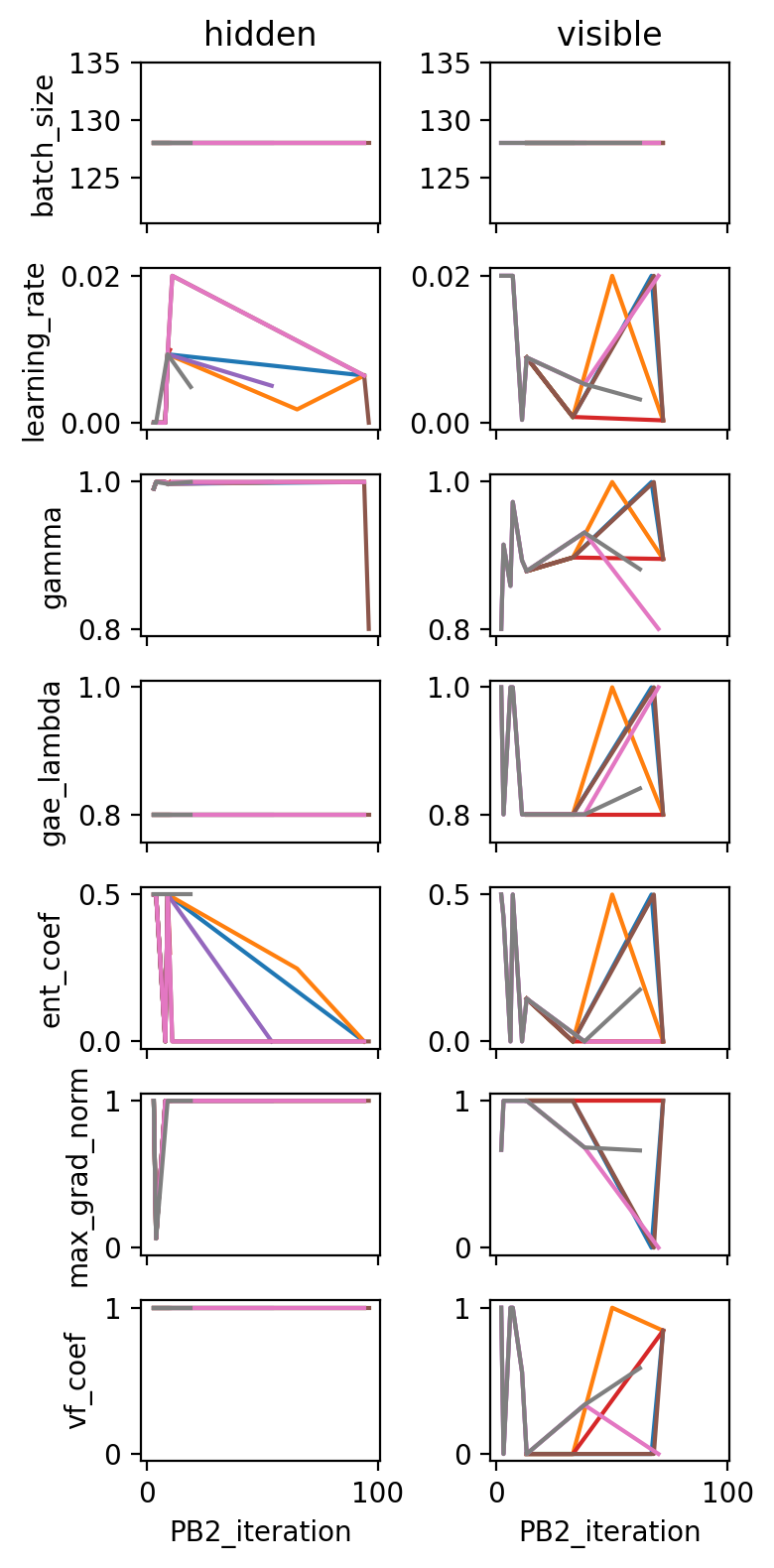}
        \label{fig:pb2_hp_schedule_CARLAcrobot}
        \caption{CARLAcrobot}
    \end{subfigure}
    \quad
    \begin{subfigure}[t]{0.3\textwidth}
        \centering
        \includegraphics[width=1\textwidth]{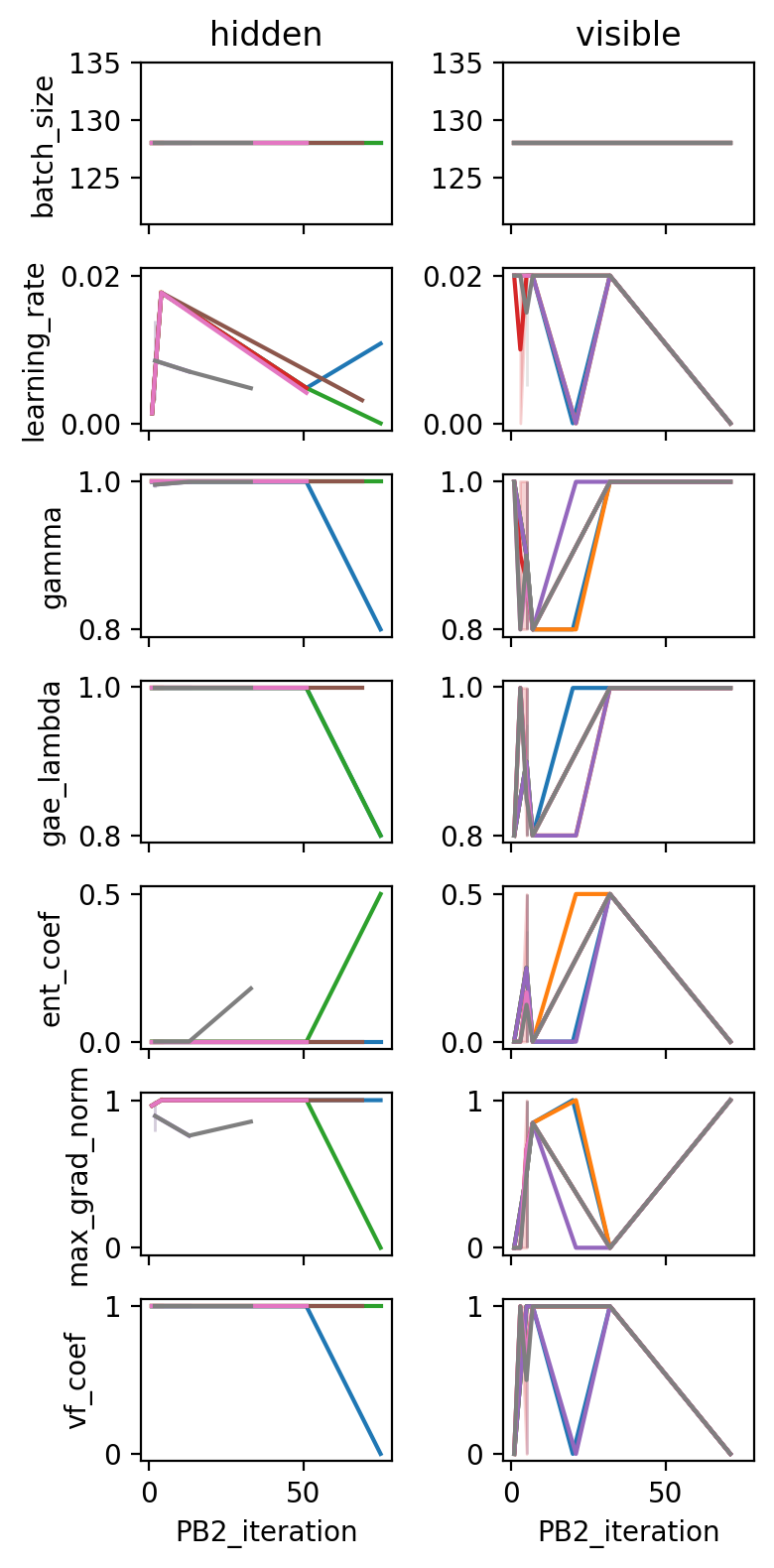}
        \label{fig:pb2_hp_schedule_CARLLunarLander}
        \caption{CARLLunarLander}
    \end{subfigure}
    \caption{Hyperparameter schedules found by PB2~\cite{parkerholder-neurips20} for hidden and visible context. The different linecolors indicate schedules found by individual PB2 workers (8 in total).}
    \label{fig:pb2_hp_schedules}
\end{figure}

The schedules found by PB2~\cite{parkerholder-neurips20} are visualised in Figure~\ref{fig:pb2_hp_schedules}.
Please note that not all workers finished the desired number of timesteps due to the memory and time limits.

\section{Context Training Distribution}
During training only one context feature is varied in each environment.
For training we use a set of 100 contexts.
We sample the context feature from a Gaussian distribution, see Table~\ref{tab:context_distributions} for details.
The default value of the context feature is used as the mean $\mu$.
\begin{table}[ht]
    \centering
    \caption{Sampling context features from a Gaussian distribution $\mathcal{N}(\mu, \sigma)$.}
    \label{tab:context_distributions}
    \begin{tabular}{lrrr}
    \toprule
         Environment & Context Feature & $\mu$ & $\sigma$ \\
         \midrule
         CARLPendulum & gravity (g) & $10$ & $0.1 \cdot 10$ \\
         CARLAcrobot & link\_length\_1 & $1$ & $0.1 \cdot 1$\\
         CARLLunarLander & gravity (GRAVITY\_Y) & $-10$ & $0.1 \cdot 10$ \\
         \bottomrule
    \end{tabular}
\end{table}

\section{Policy Sizes on CARLLunarLander}
\label{app:sizes}
\begin{figure}[ht]
    \centering
    \includegraphics[width=0.75\textwidth]{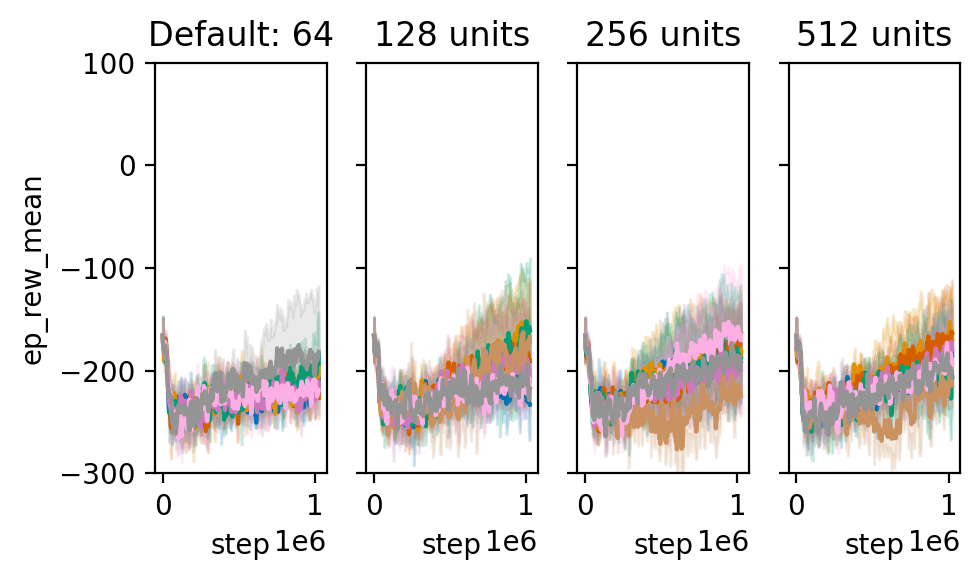}
    \caption{CARLLunarLander with increased number of policy units.}
    \label{fig:sizes}
\end{figure}

We identify the policy size as a possible cause for the poor results on CARLLunarLander.
As PB2 cannot currently optimize discrete hyperparameters, we were not able to tune it in addition to our other hyperparameters.
The default policy network has two hidden layers with $64$ units each.
Rerunning the found hyperparameter policies with a greater amount of units (see Figure~\ref{fig:sizes}) shows a slightly more positive trend than the experiment in the main paper does. 
This suggests that the architecture should be adapted in CARLLunarLander, even with only smaller variations in the gravity as used here. 
It is unclear, however, if this would solve the problem entirely, as overall performance is only increased by a small margin.
A reason might be that the network size is correlated to the other hyperparameters.

\end{document}